\documentclass[twoside]{article}
\usepackage[accepted]{aistats2021}
\usepackage{natbib}
 \pdfoutput=1
\usepackage[ruled,vlined]{algorithm2e}
\usepackage{hyperref}
\usepackage{url}
\usepackage{graphicx}
\usepackage{xcolor}
\usepackage{amssymb}

\newcommand{\fracpartial}[2]{\frac{\partial #1}{\partial  #2}}
\newcommand{\bx}{\mathbf{x}}
\newcommand{\bv}{\mathbf{v}}

\newcommand{\bz}{\mathbf{z}}

\newcommand{\by}{\mathbf{y}}

\newcommand{\cR}{\mathcal{R}}

\newcommand{\btheta}{\pmb{\theta}}
\usepackage{subfigure}
\usepackage{comment}
\usepackage[toc,page,header]{appendix}

\usepackage{multirow}
\usepackage{float}
\setcitestyle{authoryear,open={(},close={)}}

\begin{document}

%

%

\twocolumn[

\aistatstitle{Delay Differential Neural Networks}

\aistatsauthor{ Srinivas Anumasa \And P. K. Srijith }

\aistatsaddress{ Computer Science and Engineering \\ Indian Institute of Technology Hyderabad \\ \{cs16resch11004,  srijith\}@iith.ac.in } ]

\begin{abstract}
  Neural ordinary differential equations (NODEs) treat computation of intermediate feature vectors as trajectories of ordinary differential equation parameterized by a neural network. In this paper, we propose a novel model, delay differential neural networks (DDNN), inspired by delay differential equations (DDEs). The proposed model considers the derivative of the hidden feature vector as a function of the current feature vector and past feature vectors (history). The function is modelled as a neural network and consequently, it leads to continuous depth alternatives to many recent ResNet variants. We propose two different DDNN architectures, depending on the way  current and past feature vectors are considered. For training DDNNs, we provide a memory-efficient adjoint method for computing gradients and back-propagate through the network. DDNN improves the data efficiency of NODE by further reducing the number of parameters without affecting the generalization performance.  Experiments conducted on synthetic and real-world image classification datasets such as Cifar10 and Cifar100 show the effectiveness of the proposed models. 
\end{abstract}

\section{INTRODUCTION} 

The ability of deep learning models to capture rich representations of high dimensional data has led to its successful application   in computer vision problems like image classification~\citep{huang2017densely,he2016resnet}, and image captioning~\citep{Vinyals_Toshev_Bengio_Erhan_2015}.  The backbone of many such tasks is a   deep learning model called Residual Networks (ResNets)~\citep{he2016resnet}. They allowed deep learning to solve complex computer vision tasks by training deep neural networks with more than 100 layers without suffering from vanishing gradient problem. ResNets achieve this by using skip connections where input at any layer is added to the output of that layer modelling the identity mapping.    

The feature mappings in ResNet can be considered as discretization of a continuous solution modelled using  ordinary differential equations (ODE)~\citep{lu2018beyond}. 
Based on this idea, a generalization of the ResNet  was introduced, called neural ordinary differential equations (NODE)~\citep{node}. NODE is an ordinary differential equation parameterized by a neural network and can be seen to generalize ResNets to arbitrarily many layers. Neural ODE has been shown to achieve a performance close to ResNet but with a reduced number of parameters. Like in ResNets, the representations learned through NODE blocks are finally mapped to the output through a fully connected neural network (FCNN).  Neural ODEs are also shown to be more robust than convolutional neural networks (CNN)~\citep{robustode}. 

In this paper, we propose a novel continuous depth neural network model,  delay differential neural networks (DDNN), inspired by delay differential equations (DDE). The proposed model assumes the derivative of the feature vector depends not only on the current feature vector but also on the feature vectors computed in the past~\footnote{A concurrent work~\citep{anonymous2021neural}   combines neural networks and delay differential equations using a similar approach.}. 
DDNN can be considered as a generalization to NODE models  that utilize past feature vectors in addition to the present feature vector for modelling the feature mappings. Similar to NODE, the depth is not fixed but the number of feature vectors required from the past is fixed. The proposed DDNN model will improve upon the modelling capabilities of NODE, and further reduces the number of parameters in a NODE model. DDNN can be seen as the continuous depth generalization to deep neural network models like DenseNet~\citep{huang2017densely} and MixNet~\citep{wang2018mixed}. They showed that designing a neural network architecture by utilizing the previously computed feature vectors can improve generalization performance of ResNets. 
Inspired by these recent advances in ResNet variants, we consider different architectures of the DDNN model, which differs in the way historical feature vectors are combined and demonstrate their usefulness in the synthetic and image classification data sets. 

Our contributions can be summarized as follows:
\begin{itemize}
    \item We propose delay differential neural networks (DDNN), a continuous depth deep learning model based on delay differential equations parameterized by neural networks.  
    \item We propose two different DDNN architectures, considering two different approaches to combine current and past feature vectors in a DDNN model.   
    \item We  provide a memory-efficient adjoint method for updating the parameters of DDNN during backpropagation.
    \item  We conduct experiments on synthetic data sets and image classification datasets such as Cifar10 and Cifar100 datasets demonstrating the effectiveness of DDNN.
\end{itemize}
\section{RELATED WORK}
NODE~\citep{node} generalizes ResNet~\citep{he2016resnet} as continuous time representation  of discrete feature vectors obtained from a Residual block transformations. In~\citep{node}, a memory efficient adjoint method was proposed for learning parameters in a NODE model. Following this work, a series of approaches were proposed~\citep{anode,augmentedode,neuralspline} to address the learning issues in NODE. 
It was observed that generalization performance of NODE got degraded when it was made more deeper by concatenating  multiple NODE blocks.  The issue was caused due to inconsistency among the feature vectors computed during forward propagation and backpropagation which lead to incorrect computation of gradients.  Adapative checkpoint and modified adjoint approaches~\citep{ACA, anode} were proposed to address the  computation of accurate gradients. 
On the other hand, Augmented NODE~\citep{augmentedode} improves the function modelling capability of the NODE by augmenting the space (adding additional dimensions) in which NODE is solved. In contrast, we propose to consider feature representations from previous time steps to improve the function modelling capability in NODE.

 \section{BACKGROUND}
Let $\mathcal{D} =  \{X,\by\} =\{(\bx_i,y_i)\}_{i=1}^{N}$ be set of training datapoints with $\bx_i \in \cR^D$ and $y_i \in \{1,\ldots,C\}$. We denote the test data point as $(\bx_*, y_*)$. 
The aim is to learn a function which maps from input $\bx$ to a class label $y$ from the data so that it will have good generalization performance. We assume the function learnt using a neural network model to be denoted by $f$. The hidden layers in a neural network are denoted as $\bz$. In this section, we will provide background information required to understand the  proposed model. 

\subsection{ResNets and  Variants} 
\begin{figure}
    \centering
    \includegraphics[scale=0.6]{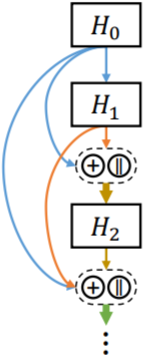}
    \caption{Path topology of MixNet~\citep{wang2018mixed}. The symbols ``$+$" and ``$||$" denote element-wise addition and
concatenation respectively.}
    \label{fig:mixnet}
\end{figure}

In standard deep learning models, increasing the  depth often hampers the performance of the model, due to vanishing gradient problem~\citep{glorot2010understanding,bengio1994learning,he2016resnet}. 
\begin{figure}
    \centering
    \includegraphics[scale=0.6]{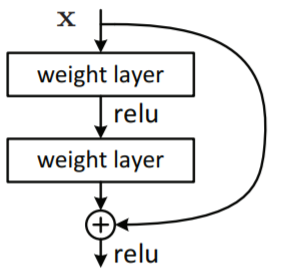}
    \caption{A residual block in ResNet architecture~\citep{he2016resnet} with skip connection}
    \label{fig:res_block}
\end{figure}
ResNet~\citep{he2016resnet} overcomes the vanishing gradient problem by allowing skip connections into the architecture as shown in the Figure~\ref{fig:res_block}.  Consequently, the function mapping for a layer $t$ is given by 
\begin{equation}
\textbf{z}_{t+1} = \textbf{z}_t + f(\textbf{z}_t,\pmb{\theta}_t)
\end{equation}
Where, $f$ is a neural network mapping parameterized by $\pmb{\theta}_t$. ResNet showed improvement in accuracy with increase in depth, where the performance of other deep learning models got degraded with increase in depth. ResNets although proven to be powerful, training time and parameter complexity has become a major hurdle for training a deeper ResNet. Stochastic depth ResNet~\cite{huang2016stochasticdepth} randomly drops subset of layers during training, resulting in reduced training time. Motivated by the success of deep learning models built on the idea of maximum information (gradients or features) flow~\citep{he2016resnet,larsson2017fractalnet}, DenseNet architecture was proposed.  DenseNet~\citep{huang2017densely}  utilizes all the computed feature in the past by concatenating across channel dimension. As there is no need to re-learn the redundant features, fewer number of parameters are required compared to ResNet.
\begin{figure}
    \centering
    \includegraphics[scale=0.6]{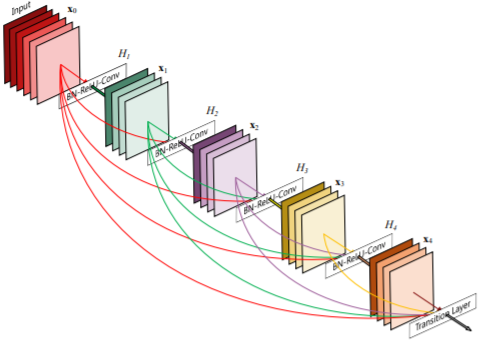}
    \caption{A 5-layer dense block~\citep{huang2017densely},
each layer takes all preceding features as input.}
    \label{fig:dense_block}
\end{figure}
ResNet~\citep{he2016resnet} and DenseNet~\citep{huang2017densely} differ in the way past feature vector are considered. While Resnet adds the past feature vectors to the current one, DenseNet concatenates them with the current one.  Mixed Link Network (MixNet)~\citep{wang2018mixed} exploits the advantages of both the models and proposed an architecture to learn better representations. They improve upon DenseNet and ResNet by using a more general architecture which considers both  ResNet(addition) and DenseNet(concatenation) operations on of features vectors (Please see Figure~\ref{fig:mixnet}).  MixNets provided improved parameter efficiency while achieving state-of-the-art accuracy on benchmark datasets.  

\subsection{Neural Ordinary Differential Equation(ODE)}

Deep learning models learn a sequence of  transformation through different spatial domain to map input $\bx_i$ to output $y_i$. In a ResNet block computation of a hidden layer representation  can be expressed using the following transformation. 
\begin{equation}
    \bz(t+1) = \bz(t) + f(t,\bz(t),\pmb{\theta}(t))
    \label{resnet}
\end{equation}
where $\textbf{z}(t)$ is feature vector with $t\in\{0...T\}$ and $f$ is neural network parameterized by parameters $\pmb{\theta}(t)$. If we use the same transformation at every step, the expression in  \eqref{resnet} is equivalent to computing the trajectory of an ordinary differential equation (ODE) using Euler method with step size one. This forms the basis of neural ODEs (NODE)~\citep{node} and it can be seen as solving an ordinary differential equation 
\begin{equation}
    \frac{d z(t)}{dt}= f(t,\textbf{z}(t),\pmb{\theta})
    \label{Eulerstep}
\end{equation}
 
Given initial feature vector $\textbf{z}(0)$ the final feature vector $\textbf{z}(T)$ can be computed by solving the ODE \eqref{Eulerstep}.  On the final feature vector $\textbf{z}(T)$, necessary transformations are applied using a fully connected neural network (FCNN), involving multiple linear mapping and activation to predict class probabilities. Parameters of the NODE model are learnt using cross-entropy loss function and memory-efficient adjoint approach to compute gradients.

\subsection{Delay Differential Equations(DDE)}
In ODEs, the derivative of the state vector with respect to time is modelled as a  function of the current state vector at that time. 
\begin{equation}
\frac{d\bz (t)}{dt} =  g(t,\bz (t))    
\end{equation}
where $\bz(t) \in \mathbb{R}^d$, $g :\mathbb{R}^{d+1}\to \mathbb{R}$. 
The solution to the ODE, i.e. the state vector value at time $t$ is obtained as a solution to the  initial value problems(IVP) (with initial state $\bz (0)$) using numerical techniques like Euler or Runga-kutta method. 

In delay differential equations (DDE), the derivative of the state vector with respect to time can be seen as a function of the current state vector and previously computed state vectors. DDEs fall under the class of infinite-dimensional dynamical systems, where the function evolves in time. 
A typical delay differential equation is given by \eqref{DDE_eq}, where, $\tau_1,\tau_2 ...\tau_n$ are constant delay values and $\phi(t)$ is the history function. 
\begin{eqnarray}
 \frac{d\bz(t)}{dt} &=&  g(t,\bz(t),\bz(t-\tau_1),...,\bz(t-\tau_n)),  \quad t > t_0 \nonumber \\
 \textbf{z}(t) &=& \phi(t), \quad  t\leq t_0
 \label{DDE_eq}
\end{eqnarray}
In this case, the change in state vector not only depends on the current state vector but also on the previous $n$ state vectors $\bz(t-\tau_1),...,\bz(t-\tau_n)$. For a DDE, due to this dependence initial value of the state $\bz(t_0)$ alone is not enough to compute the trajectory. It also requires a history function which allows it to compute the state values at time $t \leq t_0$.  In general, the delays can be a function of state vectors or can be any function of time. However, we restrict to positive valued delays. 

\begin{figure}
    \centering
    \includegraphics[scale=0.25]{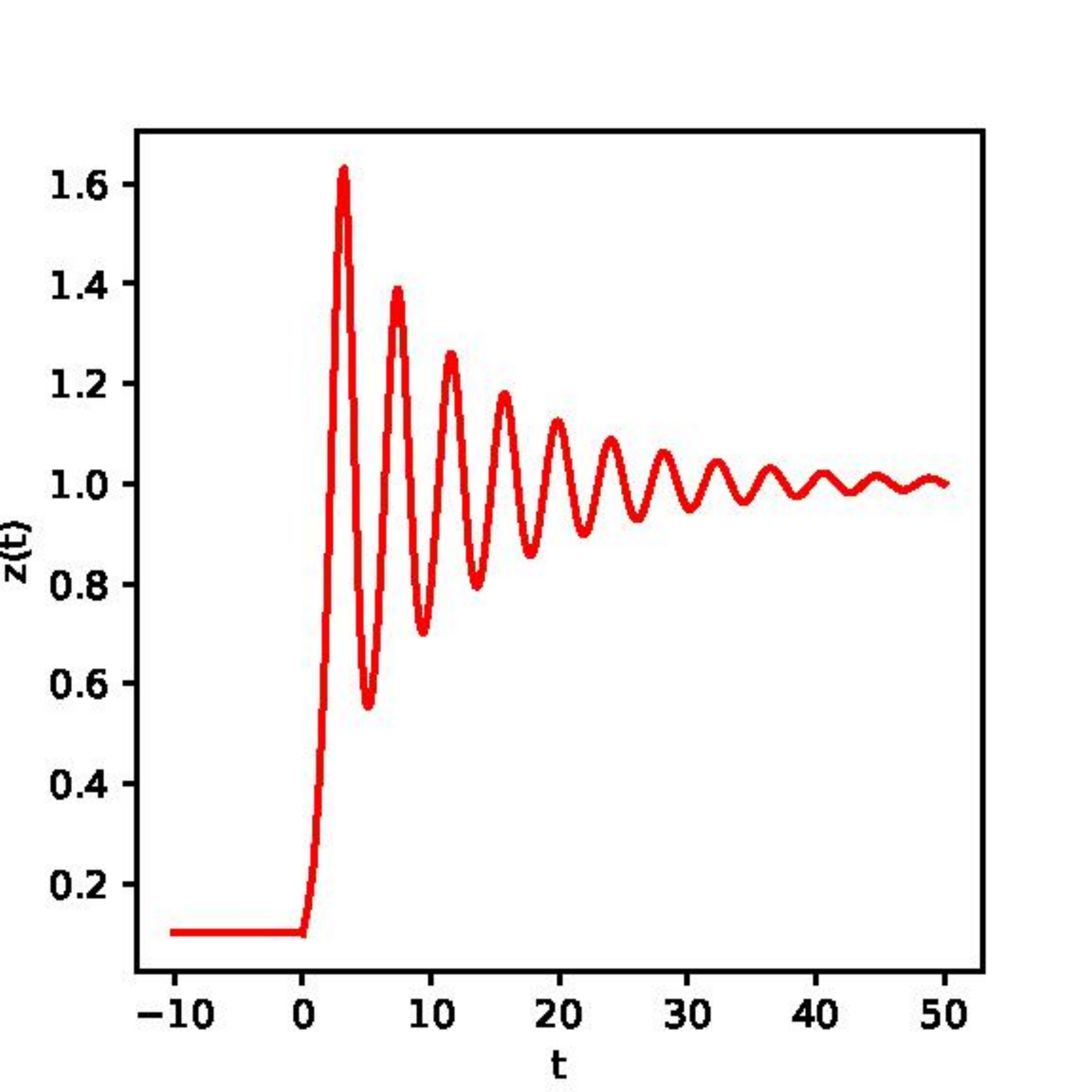}
    \caption{Solution to delay logistic equation as shown in \eqref{eq:logistic_equation}, with a =1.4 }
    \label{fig:DDE_logistic}
\end{figure}
Let us consider an example of a DDE with one constant delay. \
\begin{eqnarray}
 \frac{dz(t)}{dt} &=&  az(t)(1-z(t-1)),  \quad t > 0 \nonumber \\
 z(t) &=& 0.1, \quad  t\leq 0
 \label{eq:logistic_equation}
\end{eqnarray}

The above DDE is called delay logistic equation, famously known for its ability to capture the dynamics of population~\citep{kuang1993delay}. Figure~\ref{fig:DDE_logistic} shows the plot of the solution to \eqref{eq:logistic_equation} computed using a DDE solver.
DDEs can be solved as a sequence of IVPs for ODEs (considering the history function) and can be shown to have a unique solution in the interval of interest.  DDE with constant delays have a solution which behave differently from the ODEs. In contrast to ODEs which produce smoothly varying trajectories, DDE trajectory can exhibit abrupt changes and jumps. In  fact, one can show that DDEs  exhibit a first order discontinuity  at the initial point and higher order discontinuities in the intervals which are multiple of time delays. Hence, the abrupt changes in the DDE trajectory may get smoothen out eventually. This behaviour could be quite useful in modelling the changes in feature representations in neural networks, and motivates us to develop delay differential neural networks which we explain in detail in the following section. 

\subsection{Numerical Method for DDE}
\label{sect:num_dde}
The numerical methods to solve DDE can be obtained by extending the numerical methods to solve an ODE such as, Euler method and Runga-kutta (RK) methods~\citep{shampine1994numerical} by considering the  delay term.  
For instance, a $q^{th}$ order, fixed step size,  explicit RK method to solve a  DDE~\citep{dde_numerical} is shown as in \eqref{eq:RK}. Where, $h$ is the step size and values of the parameters $\{{b_i,c_i}\}_{i=1}^{q}$, $\{a_{ij}\}_{i=1,j=1}^{q,i-1}$ depend on the order $q$ of RK method. 
\begin{eqnarray}
 \label{eq:RK}
& & \bz(t+h) = \bz(t) + h\sum_{i=1}^{q}b_i k_i \\
& &  k_i = f(t+c_{i}h,\bz(t)+h\sum_{j=1}^{i-1}a_{ij}k_j,\bz(t+c_{i}h-\tau)) \nonumber
\end{eqnarray}

In this work, to solve a DDNN model we opted for RK12 adpative numerical method, which is a $2$ stage numerical method as shown in \eqref{eq:RK_dde}. With values, $b_1 = 0.5$, $b_2 = 0.5$, $a_{21} = 1$, $c_1 = 0$, $c_2 =1$.

\begin{eqnarray}
 \label{eq:RK_dde}
& & \bz(t+h_t) = \bz(t) + h_t(0.5k_{1}+0.5k_{2}) \\
& &  k_1 = f(t,\bz(t),\bz(t-\tau))  \nonumber \\
& &  k_2 = f(t+h_t,\bz(t)+h_{t}k_1,\bz(t+h_t-\tau)) \nonumber \\
& & error = -h_t0.5 k_1 + h_t0.5 k_2 \nonumber
\end{eqnarray}
Where, $h_t$ is the step size accepted by the solver, which depends on the computed error. For computing history term $\bz(t+h_t-\tau)$, we need to have access to the computed feature vectors in the past and their accepted time values. In our work, we choose linear interpolation \eqref{eq:linear_inter} to compute a feature vector at any time in the past.  

\begin{eqnarray}
 \label{eq:linear_inter}
 \bz(t_j) = \bz(t_i) + \frac{\bz(t_k) - \bz(t_i)}{t_k - t_i}(t_j -t_i)
\end{eqnarray}
Let, the solver requested a feature vector from the past with time $t_j$. If no feature vector is computed at time $t_j$, a pair of time points $(t_i, t_k)$ satisfying $t_i < t_j < t_k $ are chosen for computing $\bz(t_j)$ using linear interpolation \eqref{eq:linear_inter}.


\section{DELAY DIFFERENTIAL NEURAL NETWORKS}
Neural ODEs have been shown to be quite useful in many tasks which were earlier solved using ResNets. They allowed the neural networks to grow arbitrary deeper without increasing the number of parameters.  However, their generalization performance was not close to ResNets, and several approaches were proposed to improve upon it.  In this section, we propose an approach, delay differential neural networks (DDNN), which  would improve the function modelling capability of NODE and further reduces the number of parameters in the model. The proposed approach utilizes the past feature vectors to compute the next feature vector, through the framework of delay differential equations.  

\begin{figure}[t]
    \centering
    \includegraphics[scale=0.6]{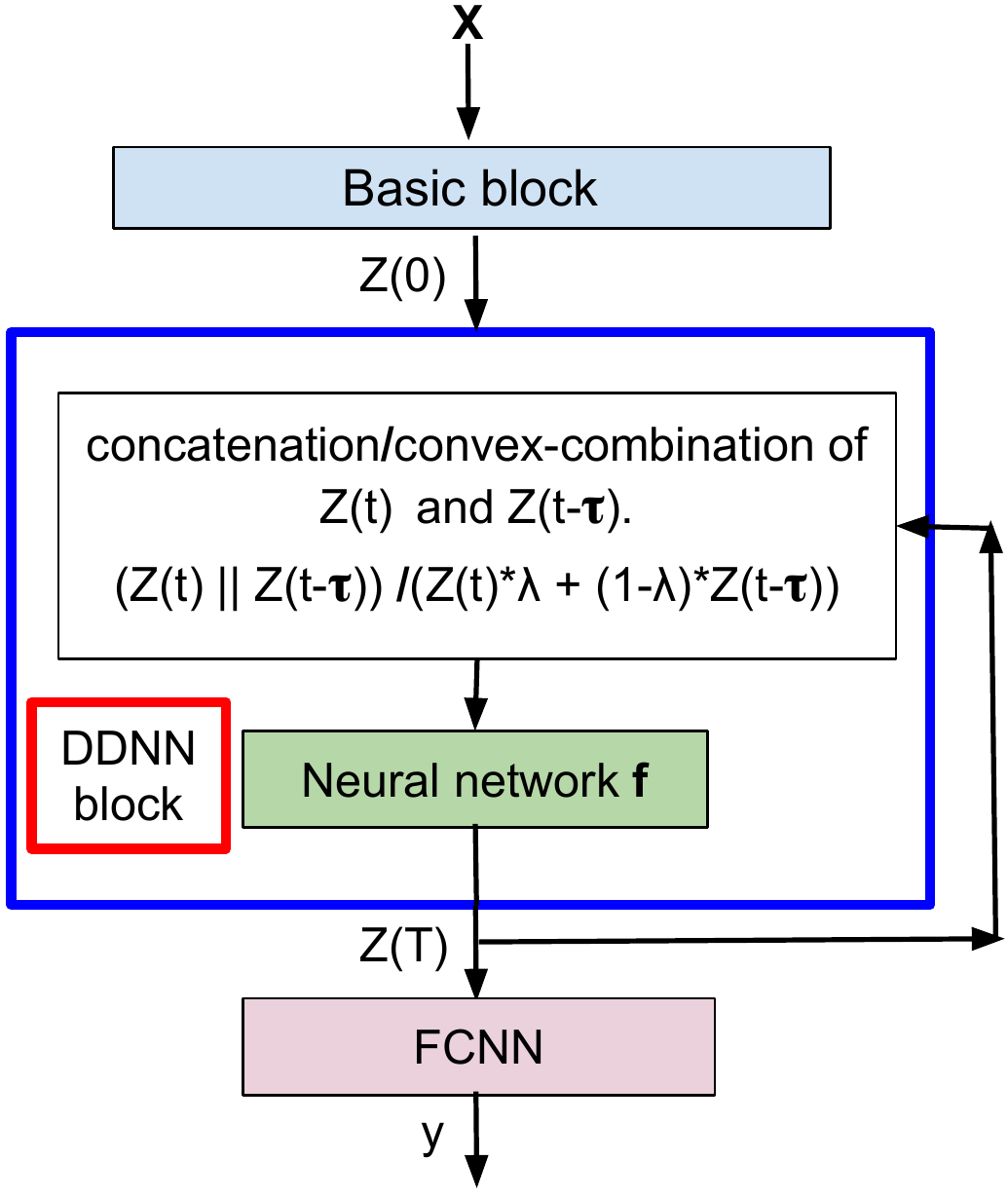}
    \caption{Feature transformations in a DDNN model considering either concatenation or combination operations over past and present feature vectors}
    \label{fig:NDDE_model}
\end{figure}

The proposed delay differential neural networks (DDNN) consider the following dynamics to model the feature representations $\bz(t)$
\begin{eqnarray}
 \frac{d\bz(t)}{dt} &=&  f(t,\bz(t),\bz(t-\tau),\pmb{\theta}) , \quad t > t_0 \nonumber, \tau > 0 \\
 \textbf{z}(t) &=& \phi(t) , \quad t\leq t_0
 \label{ours}
\end{eqnarray}
Where $f$ is a neural network parameterized by $\theta$ which consider not only the current feature vector but also  the previously computed feature vector at time $t-\tau$.  Here we restrict ourselves to a single constant delay, while the proposed framework can be generalized to multiple delays as well. The initial representation $\bz(0)$ is obtained through some basic operation (e.g. downsampling) of the  input $\bx$. This is transformed through  DDNN defined in \eqref{ours} to obtain the final representation $\bz(T)$ at some time $T$. The final representation $\bz(T)$ is transformed using a fully connected neural network to obtain the output $y$. Figure \ref{fig:NDDE_model} provides a pictorial representation of the transformations in a  DDNN model. The final representation $\bz(T)$ is obtained by solving the initial value problem defined by the DDNN \eqref{ours}. The solution to the DDNN model can be obtained using numerical techniques such as the Runga-Kutta method discussed in Section \ref{sect:num_dde}. Solving DDNN requires one to know the values of $\bz(t)$ for $t \leq t_0$, and is provided by the history function  $\phi(t)$. We consider the history function $\phi(t) = \bz(0)$ for $t \leq t_0$.  The key component in the proposed DDNN model is the function  $f$ which consider the current and past feature vectors. 
We propose two variants of the DDNN model, which differs in the way these feature vectors are used. In one model,  we consider a concatenation of the feature vectors while in the second model we consider a combination of the feature vectors.

\subsection{Concatenation of  Feature Vectors}

We first consider a DDNN model (DDNN\_Cat) which concatenates past feature vector $\bz(t-\tau)$ with the current feature vector $\bz(t)$, and provides this concatenated feature vector as an input to the function $f$. This is loosely inspired from the DenseNet architecture which does the concatenation of feature vectors.  For the DDNN\_Cat model, the size of the input and output feature vectors fed to the neural network transformation are not the same. The neural network is designed in such a way, that the incoming higher dimensional feature vector is transformed to a lower dimensional feature vector. This process is continued until the time component $t$ reaches $T$.  In our experiments for image classification, a DDNN block is constructed based on convolutional neural networks (CNN) which deal with tensors valued features with multiple channels. These features are concatenated across channel dimension and the CNN  computes a tensor of reduced channels. An advantage of the DDNN\_Cat is that it can lead to a reduced number of parameters with a competitive generalization performance compared to NODE and Resnet. With DDNN\_Cat, one can work with a half the output size compared to NODE but without affecting the performance due to its consideration of previous feature vectors. The dynamics of the DDNN\_Cat can be written as 
\begin{eqnarray}
 \frac{d\bz(t)}{dt} &=&  f(t,h(\bz(t),\bz(t-\tau)),\pmb{\theta}) ,\quad t > t_0 \nonumber \\
 \textbf{z}(t) &=& \phi(t) , \quad t\leq t_0
 \label{eq:concat}
\end{eqnarray}
where $h$ denotes an operation over the feature vectors. In the case of DDNN\_Cat, $h(\bz(t),\bz(t-\tau)) = z(t)||z(t-\tau)$. If we use Euler method to compute the trajectory of the DDE, it will take the following form,
\begin{equation}
    \bz(t+1) = \bz(t) + f(t,h(\bz(t),\bz(t-\tau)),\pmb{\theta})
\end{equation}
The parameter $\tau$ here is a hyperparameter which is determined through the  validation data. 

\begin{table*}[t]
    \caption{Comparing adjoint methods used for NODE and DDNN during Backpropagation. where $f_{\bv}(t)$ is $\fracpartial{f(t,\bz(t),\bz(t-\tau),\btheta)}{\bz(t-\tau)}$, $f_{\bz}(t)$ is $\fracpartial{f(t,\bz(t),\bz(t-\tau),\btheta)}{\bz(t)}$, $f_{\btheta}(t)$ is $\fracpartial{f(t,\bz(t),\bz(t-\tau),\btheta)}{\btheta}$ and $\bz_{\btheta}(t)$ is  $\fracpartial{\bz(t)}{\btheta}$  }
   
    \label{tab:comp_adj}
    \begin{center}
    \begin{tabular}{ |p{4cm}|p{10cm}| }
          \hline
          \hspace{1cm}\textbf{NODE} & \hspace{4cm}\textbf{DDNN}\\ 
          \hline
          \hspace{0.5cm}$\alpha^{T}(t) = - \fracpartial{L(\bz(t))}{\bz(t)}$ & \hspace{3cm} $\alpha^{T}(t) = - \fracpartial{L(\bz(t))}{\bz(t)}$  \\
          \hline
          \hspace{0.5cm}$\frac{d\alpha^{T}(t)}{dt} = -\alpha^{T}(t)f_{\bz}(t)$ & \hspace{2cm}$\frac{d\alpha^{T}(t)}{dt} = -\alpha^{T}(t)f_{\bz}(t) - \alpha^{T}(t+\tau)f_{\bv}(t+\tau)$\\
          \hline
          \hspace{0.5cm}$\frac{dL}{d\btheta} = \int_{0}^{T}\alpha^{T}(t)f_{\btheta}(t)dt$ & \hspace{1cm}$\frac{dL}{d\btheta} = \int_{0}^{T}\alpha^{T}(t)f_{\btheta}(t)dt - \int_{-\tau}^{0}\alpha^{T}(t+\tau) f_{\bv}(t+\tau)  \bz_{\btheta}(t)dt$ \\
          \hline
    \end{tabular}
    \end{center}
\end{table*}

\subsection{Convex Combination of Feature Vectors}
The second DDNN model we consider is DDNN\_Conv, which considers  a convex combination of the current feature vector and the past feature vector in the neural network transformation in the DDNN block. In particular, it considers the operation over feature vectors as follows 
\begin{equation}
    h(\bz(t),\bz(t-\tau)) = \lambda \bz(t) + (1-\lambda) \bz(t-\tau) 
    \label{eq:linear}
\end{equation}
where, $\lambda$ and  $\tau$ are hyperparameters which can be determined based on validation data.  Note that in this case, the input and output dimension of  the neural network function in the DDNN block is the same unlike the DDNN\_Cat model and the transformations take this into account.

\section{BACKPROPAGATION USING ADJOINT METHOD} 
    We discuss an approach to learn the parameters  in the DDNN model. The parameters associated with the neural networks of various DDNN blocks can be learnt through back-propagation. To apply back-propagation standard loss functions for regression and classification can be used. The output $\hat{y}$ obtained from the FCNN block is used to compute the loss function $L$. The  loss function  can be taken as the least squares loss for regression problems  and cross-entropy loss for classification problems, for e.g. $L(y,\hat{y}) = - y \log p(\hat{y}) - (1-y) \log 1 - p(\hat{y})$. However, applying standard back propagation to learn the DDNN parameters, requires computing and  storing the gradients with respect to every state and is not memory efficient.  We use the adjoint method, a memory-efficient way of computing the gradients, for learning the parameters in the DDNN model.  This is similar to the adjoint method used in the  NODE~\citep{node,anode,ACA} models, but we derive it to the case with delay differential equations. 

\begin{algorithm}[t]
\SetAlgoLined
Initialize $\alpha^T(T),\frac{dL}{d\btheta} = 0$\\
 \textbf{For} $N$ to 1:\\
\hspace{1.0cm} (1) Compute the value of the function $f$ at $t_i$ and $t_i + \tau$ for computing $f_\bz(t_i)$,$f_{\btheta}(t)$ and $f_\bv(t_i+\tau)$. \\
\hspace{1.0cm} (2) local backward, update $\alpha(t_i)$ and $\frac{dL}{d\btheta}$ according to discretization of \eqref{eq:alpha} and~\ref{eq:theta}.
 \caption{Adjoint method during backpropagation for DDNN model}
 \label{alg:adjoint}
\end{algorithm}

\begin{figure*}[t]
\begin{center}     
\subfigure[DDNN\_Conv]{\includegraphics[scale=0.3]{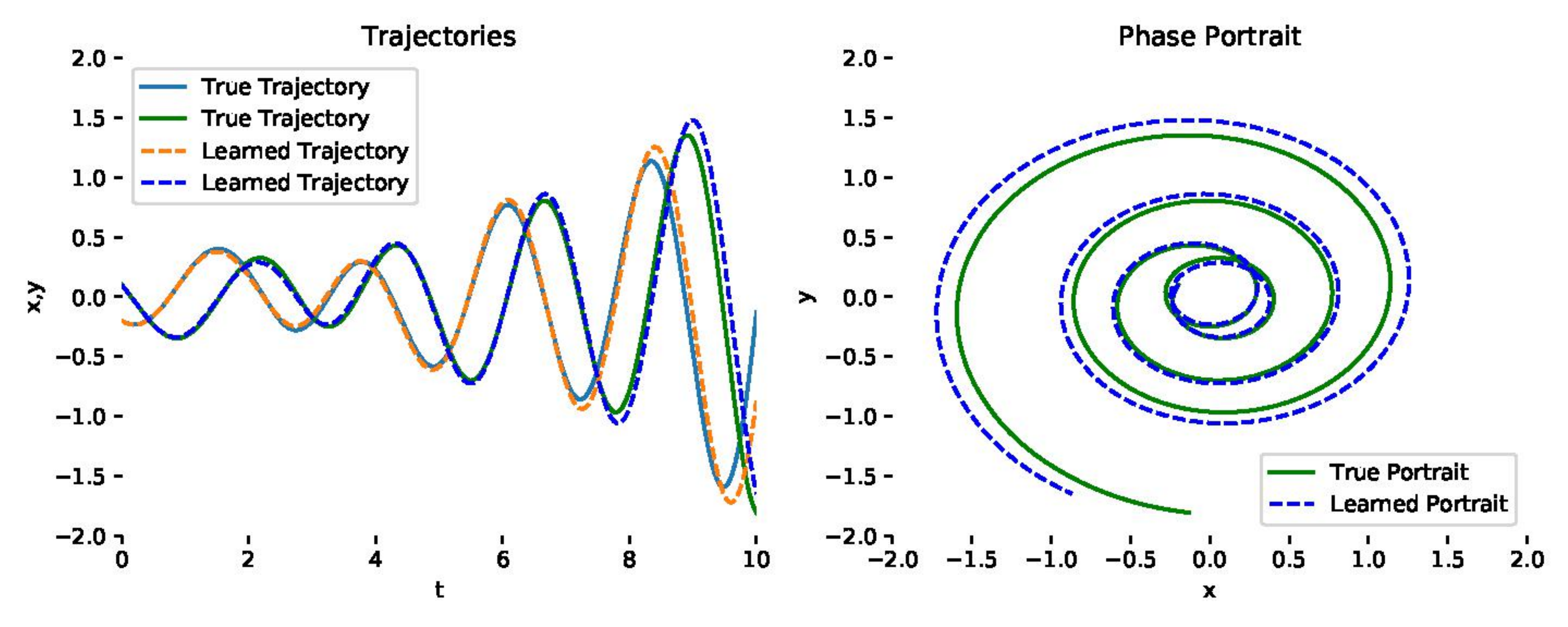}}
\subfigure[ACA\_NODE]{\includegraphics[scale=0.3]{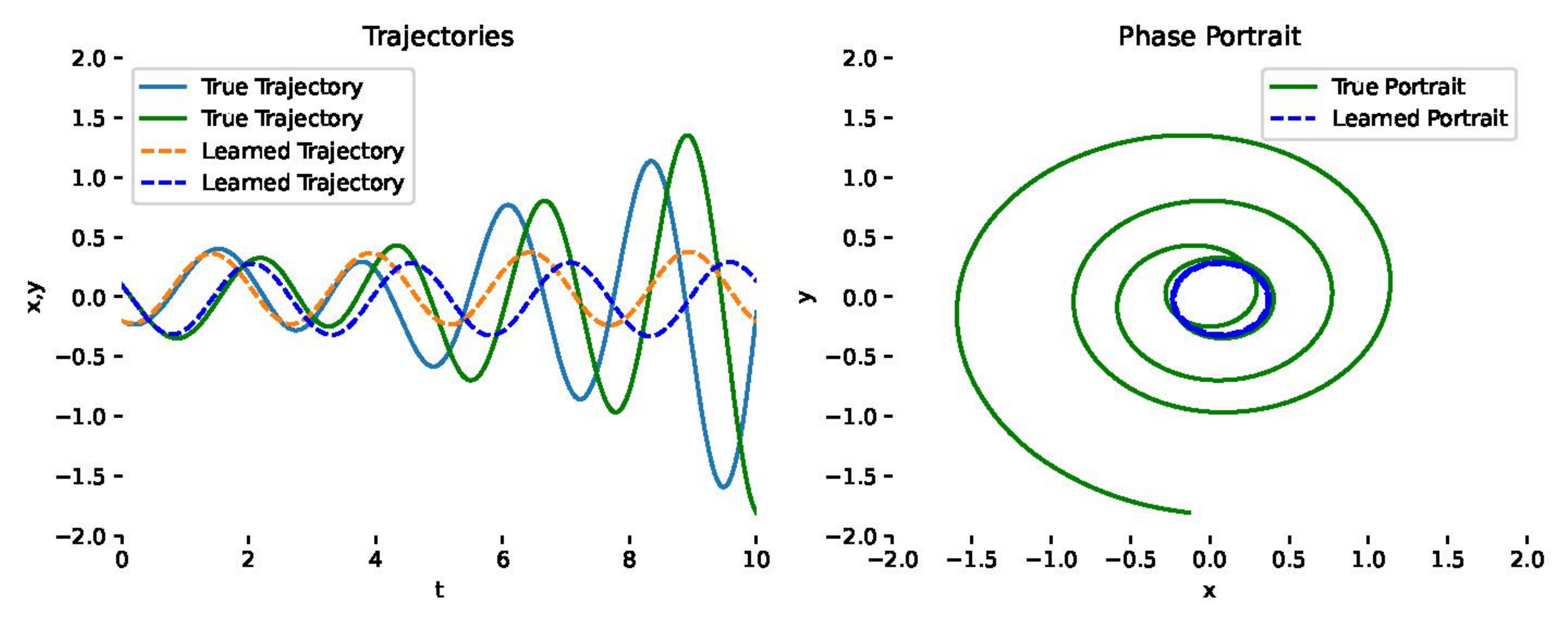}}
\end{center}
\caption{Showing the learning and extrapolating capabilities of (a)DDNN\_Conv and (b)ACA\_NODE, given samples from the solution of DDE \eqref{eq:toy_add}. Plots in the left half of both (a) and (b), shows learned and actual trajectories of individual components with respect to time. Plots in the right half of both (a) and (b) represent phase portraits of both learned and actual solution of DDE \eqref{eq:toy_add}. This is obtained by plotting the individual components against each other over time $t$, where, $x = \bz(t)[0]$ and $y = \bz(t)[1]$.  } 
\label{fig:toy_data}
\end{figure*}

For the DDNN model,  we extend the adjoint method  in~\citep{adjoint_dde} for learning the parameters of a DDE. Algorithm~\ref{alg:adjoint} provides a snippet of adjoint method used for the DDNN model and provide the detailed derivation in the appendix. The adjoint $\alpha^{T}(t)$ models the derivative of the loss function  with respect to some feature vector $\bz(t)$ at time $t$
\begin{equation}
 \alpha^{T}(t) = - \fracpartial{L(y, \hat{y})}{\bz(t)}
\end{equation} 
The adjoint can be used to compute the gradient of parameters $\btheta$ with respect to loss $L$. We can model the  dynamics of the adjoint $\alpha^{T}(t)$ over time and is given by the following delay differential equation  
\begin{equation}
\frac{d\alpha^{T}(t)}{dt} = -\alpha^{T}(t)f_{\bz}(t) - \alpha^{T}(t+\tau)f_{\bv}(t+\tau)
\label{eq:alpha}
\end{equation}
where $f_{\bv}(t)$ is $\fracpartial{f(t,\bz(t),\bz(t-\tau),\btheta)}{\bz(t-\tau)}$, $f_{\bz}(t)$ is $\fracpartial{f(t,\bz(t),\bz(t-\tau),\btheta)}{\bz(t)}$. The value of the adjoint $\alpha^{T}(t)$ at time $t$ can be found by solving  the delay differential equation \eqref{eq:alpha} backwards in time. The numerical methods for DDE based on Runga-Kutta methods discussed in Section \ref{sect:num_dde}  can be used to solve \eqref{eq:alpha}.  Computing the gradients of the parameters $\btheta$ with respect to loss $L$ requires evaluating an integral, which depends on $\bz(t)$ and  $\alpha^{T}(t)$. 
\begin{equation}
\frac{dL}{d\btheta} = \int_{0}^{T}\alpha^{T}(t)f_{\btheta}(t)dt - \int_{-\tau}^{0}\alpha^{T}(t+\tau) f_{\bv}(t+\tau)  \bz_{\btheta}(t)dt
\label{eq:theta}
\end{equation}
Where $f_{\btheta}(t)$ is $\fracpartial{f(t,\bz(t),\bz(t-\tau),\btheta)}{\btheta}$ and $\bz_{\btheta}(t)$ is  $\fracpartial{\bz(t)}{\btheta}$. 
Table~\ref{tab:comp_adj} provides a comparison of adjoint methods used for training NODE and the proposed DDNN model. 

Our approach uses adjoint method for computing gradients during backpropagation and uses adaptive checkpoint~\citep{ACA} framework for computing accurate gradients. 
The adaptive checkpoint approach requires storing the feature vectors computed during forward propagation and is then utilized during backpropagation. This is shown to be numerically more accurate than naive adjoint method~\citep{node}.  
Assuming  $N$ to be the number of states evaluated, the approach requires a memory complexity of $O(N)$ for each DDNN block.


\begin{figure}[t]
\begin{center}     
\includegraphics[scale=0.25]{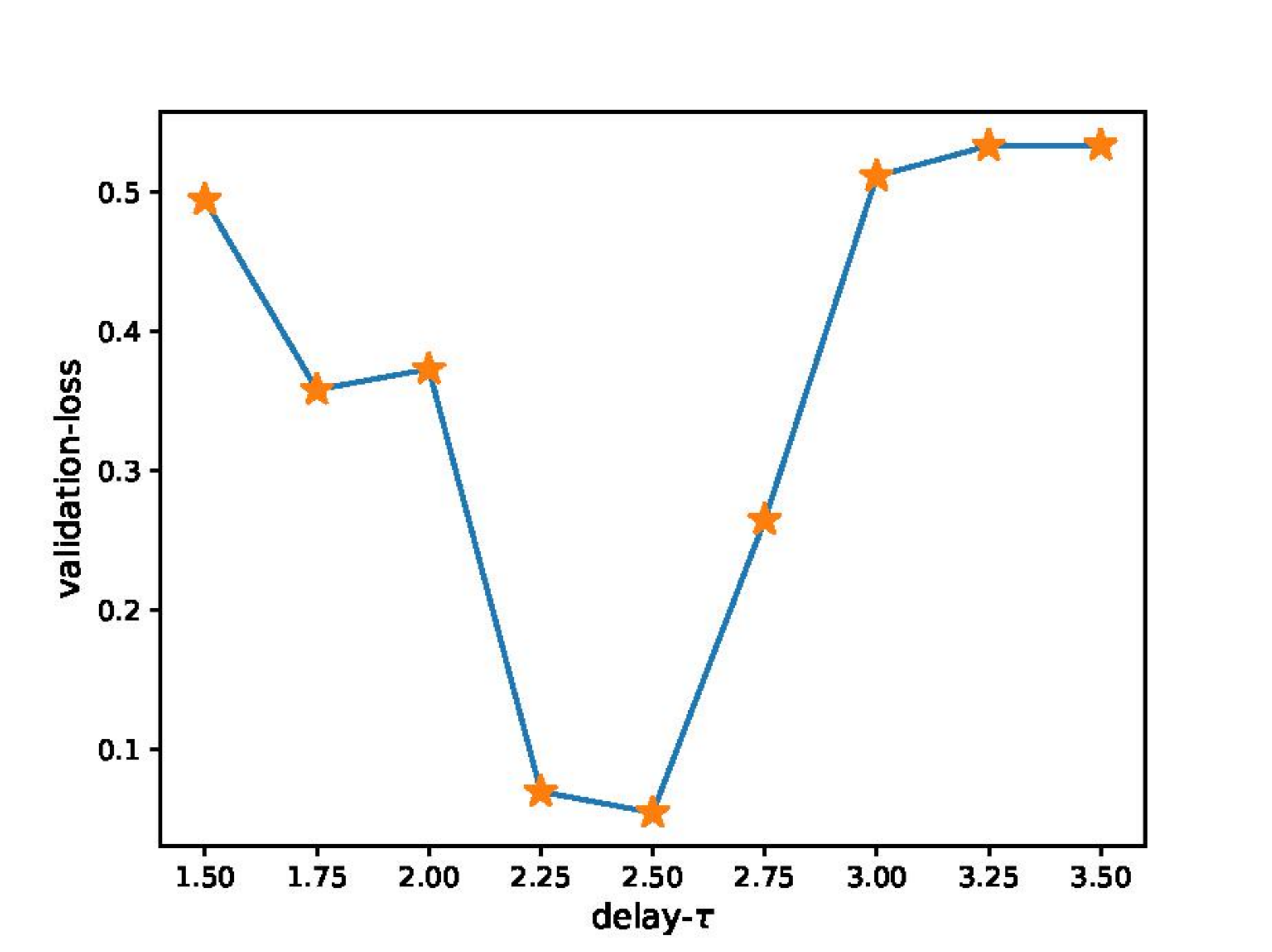}
\end{center}
\caption{Validation loss of DDNN\_Conv model with different delay $\tau$ values trained over synthetic data.}
\label{fig:toy_loss}
\end{figure}

\section{EXPERIMENTS}
We conduct experiments to study the performance of the proposed models on synthetic and real world image datasets.  
In the experiments, we compare against the state-of-the-art NODE model~\citep{ACA} model.
We conduct experiments on a toy dataset to compare the learning capability of DDNN over ACA\_NODE~\citep{ACA}. 
We also conduct experiments for image classification on the real datasets such as Cifar10 and  Cifar100, and compare against ACA\_NODE and ResNet18 models. 

\subsection{Synthetic Data experiments}
In this section we demonstrate the performance of DDNN\_Conv on a 2-D toy dataset. The synthetic data is generated from a 2-dimensional DDE \eqref{eq:toy_add}. 
\begin{eqnarray}
 \frac{d\bz(t)}{dt} &=&  0.75\bz(t) + 0.25\bz(t-2.5)\times A , t > t_0 \nonumber \\
 A &=& \begin{bmatrix}
-0.1 & 3.2 \\
-3.2 & -0.1
\end{bmatrix} \nonumber \\
 \textbf{z}(t) &=& [-0.2,0.1] ,\quad t\leq t_0
 \label{eq:toy_add}
\end{eqnarray}
We sampled a set of 1000 samples from the solution of the DDE  using a scipy-based DDE solver\footnote{https://pypi.org/project/ddeint/} available in python. The 1000 samples are sampled on a time scale $0$ to $10$ as shown in Figure~\ref{fig:toy_data}. Samples from time scale $0$ to $6$ were chosen as training samples, time scale $6$ to $8$ as validation samples. Samples from time scale $8$ to $10$ as test samples. A two layer neural network with hidden dimension 50 is used for both the ACA\_NODE and our proposed DDNN\_Conv model with $\lambda=0.75$. Mean-square error loss is computed between computed trajectory and the actual trajectory, which will be used to update the parameters of the model. 

While training DDNN\_Conv we chose $\tau$ to be a hyperparameter, and varied the value from $1.5$ to $3.5$ with an interval of $0.25$. Based on the validation loss we choose the best  $\tau$ value. Figure~\ref{fig:toy_loss} shows the performance of DDNN\_Conv models using different delay($\tau$) values. One can observe the delay with value $2.5$ is  having the lowest validation loss compared to other delay values which corroborates well with the delay used in the DDE \eqref{eq:toy_add}.  Figure~\ref{fig:toy_data}(a) shows the reconstruction capability of DDNN model with a delay $2.5$ which is having the lowest validation loss among DDNN\_Conv models with other delay values. Figures showing the learning capabilities of DDNN\_Conv models with other delay values can be found in the appendix. Although DDNN\_Conv model has seen only the samples on a time scale $0$ to $6$, it was able to generate well the samples on a time scale $6$ to $10$. Figure~\ref{fig:toy_data}(b) shows the learning and reconstruction capability of ACA\_NODE model,  which fails to learn. From this experiment, we conclude that DDNN\_Conv model able to learn the underlying dynamics provided samples from the solution of a DDE, where ACA\_NODE fails.  
\begin{table*}
 \caption{Validation accuracy of DDNN models, under different operations, convex combination and concatenation with different delay values. All the models are trained for 90 epochs}
     \label{tab:val_acc_NDDE}
     \begin{center}

\begin{tabular}{|p{2.5cm}|p{1.8cm}|*{10}{p{7mm}|}}
\hline
\textbf{MODELS}&\textbf{DATASET}&\multicolumn{10}{c|}{\textbf{DELAY}}\\
\cline{3-12}
&&\textbf{0.1}&\textbf{0.2}&\textbf{0.3}&\textbf{0.4}&\textbf{0.5}&\textbf{0.6}&\textbf{0.7}&\textbf{0.8}&\textbf{0.9}&\textbf{1.0}\\
\hline
\textbf{DDNN\_Conv}&\textbf{Cifar10}&93.50&93.62&93.60&93.58&93.30&93.74&93.62&93.78&93.82&\textbf{93.98}\\
\cline{2-12}
&\textbf{Cifar100}&74.56&73.50&73.62&74.42&74.72&74.74&73.14&\textbf{76.00}&74.86&74.72\\
\hline
\textbf{DDNN\_Cat}&\textbf{Cifar10}&92.98&\textbf{93.44}&92.96&93.02&93.22&92.88&92.94&92.88&93.18&93.24\\
\cline{2-12}
&\textbf{Cifar100}&72.20&72.14&71.92&72.18&72.84&\textbf{73.26}&71.32&72.52&72.58&72.94\\
\hline

\end{tabular}
\end{center}
\end{table*}
\begin{table*}
     \caption{Comparing test accuracy of ACA\_NODE, ResNet18 models against DDNN\_Cat and DDNN\_Conv models. For the columns DDNN\_Cat and DDNN\_Conv delay values of the best models based on test accuracy are also given.}
     \label{tab:test_acc}
     \begin{center}
        \begin{tabular}{|p{2cm}|p{2cm}|p{3cm}|p{3cm}|p{2cm}|}
            \hline
            \textbf{DATASET} & \textbf{ACA\_NODE} & \textbf{DDNN\_Conv} & \textbf{DDNN\_Cat} & \textbf{RESNET18} \\
            \hline
            \textbf{Cifar10} & 94.16 & 93.96(delay 1.0) & \textbf{ 94.30(delay 0.2)}& 94.28\\
            \hline
             \textbf{Cifar100} & 74.60 & \textbf{76.52(delay 0.8)} & 74.34(delay 0.6) & 76.08\\
             \hline
        \end{tabular}
        \end{center}
\end{table*}

\subsection{Image Classification}
Under image classification, we conducted experiments on Cifar10 and  Cifar100 datasets. Cross-entropy loss is used to compute gradients with respect to the parameters.  We split the dataset containing 60000 samples into 50000 as training, 5000 as validation data, and 5000 as test data. For DDNN models, validation data will be used to select the best delay hyperparameter. For training the models we follow a similar setup as discussed in~\citep{ACA}, where all the models are trained for 90 epochs on both Cifar10 and Cifar100. Scheduled decay learning rate is used with an initial learning rate as $0.1$ with decay as $0.1$.  The learning rate is rescheduled at epoch 30 and 60.

\subsubsection{DDNN Convex Combination}
We first study the performance of the DDNN model (DDNN\_Conv) which consider a convex combination of feature vectors. 
The parameter $\lambda$ was fixed to $0.75$ after analysing the performance on the validation data. This essentially means that we give a higher importance to the current feature vector and a lower importance to the previous feature vector.
As we stated earlier, the delay($\tau$) value is treated as a hyperparameter, 10 different DDNN\_Conv models were trained with different $\tau$ values, ranging from $0.1$ to $1.0$ with an interval of $0.1$. The architecture design and the number of parameters is same as of ACA\_NODE. 

In Table~\ref{tab:val_acc_NDDE}, the validation accuracies of DDNN\_Conv models with different delay ($\tau$) values are provided. As can be observed from   Table~\ref{tab:val_acc_NDDE}, the best  DDNN\_Conv model  for Cifar10 dataset is the one with delay $1.0$ and for Cifar100  is one with the delay $0.8$. Thus, the DDNN\_Conv model with the best validation accuracy is found to be the one with higher delay value.  The possible explanation could be, re-utilizing the older features (higher delay values) compared to re-utilizing the recently computed features (low delay values) may be helping  in  better accuracy. From Table~\ref{tab:test_acc}, one can observe that the test accuracy  of the best DDNN\_Conv model for Cifar10 is  $93.96\%$  and for Cifar100 is $76.52\%$. On Cifar100 dataset, DDNN\_Conv model achieved a higher($76.52\%$) accuracy compared to  ACA\_NODE($74.60\%$) and ResNet18($76.08\%$) models, showing the effectiveness of DDE based deep leaning models.

\subsubsection{Concatenation} 
The architecture design of DDNN\_Cat is different when compared with DDNN\_Conv and ACA\_NODE. In DDNN\_Conv or ACA\_NODE models, dimension of  the expected input feature vector (Tensor) to DDNN block or ACA\_NODE block is same as the computed feature vector (Tensor) dimension. But for DDNN\_Cat,  the input tensor and output tensor differ in terms of number of channels. 
Let $C$ be expected number of channels by both the models DDNN\_Conv and DDNN\_Cat. But, the computed tensor is different in terms  of number of channels. DDNN\_Cat produces a  tensor with $C/2$  number of  channels unlike  $C$ for DDNN\_Conv. This results in a reduced number of parameters for DDNN\_Cat ($7$ Million) compared to DDNN\_Conv  and ACA\_NODE models ($11$  Million). 

We trained the DDNN\_Cat model with different delay values and compared their validation performance in Table~\ref{tab:val_acc_NDDE}.
Test accuracy of the best performing  DDNN\_Cat model trained on Cifar10 and Cifar100 datasets are given in Table~\ref{tab:test_acc}. We can find that the DDNN\_Cat model We found that the DDNN\_Cat model is very competitive and even performed better than other models in the Cifar10 dataset, achieving an accuracy of $94.30$ for a delay value of $0.2$. Thus, DDNN\_Cat model gave a very good performance in spite of reduction in the number of parameters. 
\section{CONCLUSION}
In this work we proposed a novel continuous depth deep learning model, delay differential neural networks (DDNN), based on the  principles of delay differential equations. We provided two different DDNN architectures, DDNN\_cat and DDNN\_Conv based on the way DDNN block consider the present and past feature vectors. We also discussed an adjoint method which provides a   memory efficient way to learn parameters in DDNN. Through experiments on toy-dataset we showed that DDNN model can learn the trajectory of the solution to a DDE, where ACA\_NODE fails. We discussed the performance of DDNN models on image classification datasets such as Cifar10 and Cifar100. We showed that under concatenation operation, DDNN model with reduced number of parameters, performs well without  affecting the generalization performance. As a future work, we develop more general DDNN models and apply to other problems in computer vision.

\bibliography{DDNN}
\clearpage
\onecolumn
\begin{appendix}

\section{ADJOINT METHOD}
In this section we give the derivation of the adjoint method based on~\cite{adjoint_dde}, which is used to update the parameters of the DDNN during backpropagation. The derivations provided are for under one constant delay and this can be easily extended to multiple constant delays. 
Given DDE with a history function
\begin{eqnarray}
 \frac{d\bz(t)}{dt} =  f(\bz(t),\bz(t-\tau),t,\pmb{\theta}) , t > t_0 \nonumber \\
 \textbf{z}_{t} = \phi(t) , t\leq t_0
 \label{ours1}
\end{eqnarray}
$f$ is represented by a neural network which is differentiable almost everywhere. The forward pass is defined as:
\begin{equation}
    \hat{y} = \bz(T) = \bz(0) + \int_{0}^{T}f(\bz(t),\bz(t-\tau_1),t,\pmb{\theta})dt
\end{equation}

The loss function is defined as: 
\begin{equation}
    J(\hat{y},y) = J(\bz(T),y))
\end{equation}

To minimize the loss function we need to compute the $\frac{dJ(\hat{y},y)}{d\btheta}$. 
 Let $\alpha^T(t)$ be any vector valued function defined for $t \in [0,T]$, then the lagrangian is, 
\begin{equation}
L = J(\bz(T),\btheta)+\int_{0}^{T}\alpha^{T}(t)[\frac{\bz(t,\btheta)}{dt}-f(t,\bz(t,\btheta),\bz(t-\tau,\btheta),\btheta)]dt
\end{equation}

For ease of notation we write,
\begin{align}
\fracpartial{f(t,\bz(t,\btheta),\bz(t-\tau,\btheta),\btheta)}{\btheta} \hspace{0.2cm} &\text{as}\hspace{0.2cm} f_{\btheta}(t),  
\fracpartial{f(t,\bz(t,\btheta,\bz(t-\tau,\btheta),\btheta)}{\bz(T,\btheta)} \hspace{0.2cm}\text{as}\hspace{0.2cm} f_{\bz}(t), 
\fracpartial{f(t,\bz(t,\btheta,\bz(t-\tau,\btheta),\btheta)}{\bz(t-\tau)} \hspace{0.2cm}\text{as}\hspace{0.2cm} f_{\bv}(t),\nonumber \\ 
\fracpartial{\bz(t,\theta)}{\btheta} \hspace{0.2cm} &\text{as}\hspace{0.2cm} \bz_{\btheta}(t),
\fracpartial{\bz(t-\tau,\theta)}{\btheta} \hspace{0.2cm}\text{as}\hspace{0.2cm} \bz_{\btheta}(t-\tau)
\end{align}

taking derivative w.r.t to $\btheta$

\begin{equation}
 \frac{dL}{d\btheta} = \fracpartial{L(\bz(T,\theta))}{\bz(T,\theta)}\bz_{\btheta}(T) +\underbrace{ \int_{0}^{T} \alpha^{T}(t) (\frac{d(\bz_{\btheta}(t))}{dt})dt}_\text{ applying integration by parts} 
  -\int_{0}^{T} \alpha^{T}(t) [f_{\btheta}(t) + f_{\bz}(t)\bz_{\btheta}(t) + f_{\bv}(t)\bz_{\btheta}(t-\tau)]dt
  \label{main}
\end{equation}

\begin{equation}
\int_{0}^{T} \alpha^{T}(t) (\frac{d(\bz_{\btheta}(t))}{dt})dt =  [\alpha^{T}(t)\bz_{\btheta}(t)]^{T}_{0} - \int_{0}^{T} \alpha^{T}(t) [ \frac{d\alpha^{T}(t)}{dt}\bz_{\btheta}(t)]
\label{int_by}
\end{equation}
$\bz(0)$ is initial state doesnt depend on $\btheta$, so $\bz_{\btheta}(0)$ is $0$.

\begin{equation}
 = \underbrace{\fracpartial{L(\bz(T,\theta))}{\bz(T,\theta)}\bz_{\btheta}(T) + \alpha^{T}(T)\bz_{\btheta}(T)}_\text{Equating this part to 0}-\int_{0}^{T}\alpha^{T}(t) [f_{\btheta}(t) + f_{\bz}(t)   \bz_{\btheta}(t) + f_{\bv}(t)\bz_{\btheta}(t-\tau)]dt
\end{equation}

\begin{equation}
\left[\fracpartial{L(\bz(T,\theta))}{\bz(T,\theta)} + \alpha^{T}(t)\right] \bz_{\btheta}(T) = 0
\end{equation}

\begin{equation}
 \boxed{
 \alpha^{T}(t) = - \fracpartial{L(\bz(T,\theta))}{\bz(T,\theta)}}
\end{equation}
This choice for $\alpha^{T}(T)$ eliminates the integral involving $\bz_{\btheta}(T)$

\begin{equation}
 = 
\int_{0}^{T}\frac{d\alpha^{T}(t)}{dt} \bz_{\btheta}(t)dt-\int_{0}^{T} \alpha^{T}(t)[f_{\btheta}(t)+ f_{\bz}(t)   \bz_{\btheta}(t) ] -\underbrace{\int_{0}^{T}\alpha^{T}(t)f_{\bv}(t)   \bz_{\btheta}(t-\tau)dt}_\text{Apply change of variables}
\label{sub_here}
\end{equation}

\begin{equation}
\int_{0}^{T}\alpha^{T}(t)f_{\bv}(t)   \bz_{\btheta}(t-\tau)dt = \int_{-\tau}^{T-\tau}\alpha^{T}(t+\tau)f_{\bv}(t+\tau)  \bz_{\btheta}(t)]dt
\end{equation}
After change of variables

\begin{equation}
=\int_{0}^{T}\alpha^{T}(t+\tau) f_{\bv}(t+\tau)  \bz_{\btheta}(t)dt +
\int_{-\tau}^{0}\alpha^{T}(t+\tau) f_{\bv}(t+\tau)  \bz_{\btheta}(t)dt ]
\label{change_of_var}
\end{equation}

After substituting Equation \ref{change_of_var} in \ref{sub_here}. 

\begin{equation}
\begin{split}
\fracpartial{L}{\theta} = & \int_{0}^{T}\frac{d\alpha^{T}(t)}{dt} \bz_{\btheta}(t)dt-\int_{0}^{T} \alpha^{T}(t)[f_{\btheta}(t)+ f_{\bz}(t)   \bz_{\btheta}(t) ]  \\
& - \int_{0}^{T}\alpha^{T}(t+\tau) f_{\bv}(t+\tau)  \bz_{\btheta}(t)dt -
\int_{-\tau}^{0}\alpha^{T}(t+\tau) f_{\bv}(t+\tau)  \bz_{\btheta}(t)dt 
\end{split}
\end{equation}
After rearranging the terms.
\begin{equation}
\begin{split}
 & = \int_{0}^{T}\left[ -\frac{d\alpha^{T}(t)}{dt} - \alpha^{T}(t) f_{\bz}(t) - \alpha^{T}(t+\tau) f_{\bv}(t+\tau)\right]\bz_{\btheta}(t)dt - \int_{0}^{T} \alpha^{T}(t)f_{\btheta}(t)dt \\
& \hspace{2cm} \int_{-\tau}^{0}\alpha^{T}(t+\tau) f_{\bv}(t+\tau)  \bz_{\btheta}(t)dt
\end{split}
\end{equation}
The dynamics of the adjoint $\alpha^{T}(t)$ given by the differential equation. 

\begin{equation}
\boxed{\frac{d\alpha^{T}(t)}{dt} = -\alpha^{T}(t)f_{\bz}(t) - \alpha^{T}(t+\tau)f_{\bv}(t+\tau)}
\end{equation}

Computing the gradients with respect to the  parameters $\btheta$ requires evaluating a integral, which depends on $\bz(t)$ and  $\alpha^{T}(t)$. 

\begin{equation}
\boxed{\frac{dL}{d\btheta} = \int_{0}^{T}\alpha^{T}(t)f_{\btheta}(t)dt - \int_{-\tau}^{0}\alpha^{T}(t+\tau) f_{\bv}(t+\tau)  \bz_{\btheta}(t)dt}
\end{equation}

\section{PERFORMANCE OF DDNN MODELS ON TOY-DATASET USING DIFFERENT DELAY VALUES}

\begin{figure}[H]
\begin{center}     
\subfigure[DDNN with delay 1.5]{\includegraphics[scale=0.15]{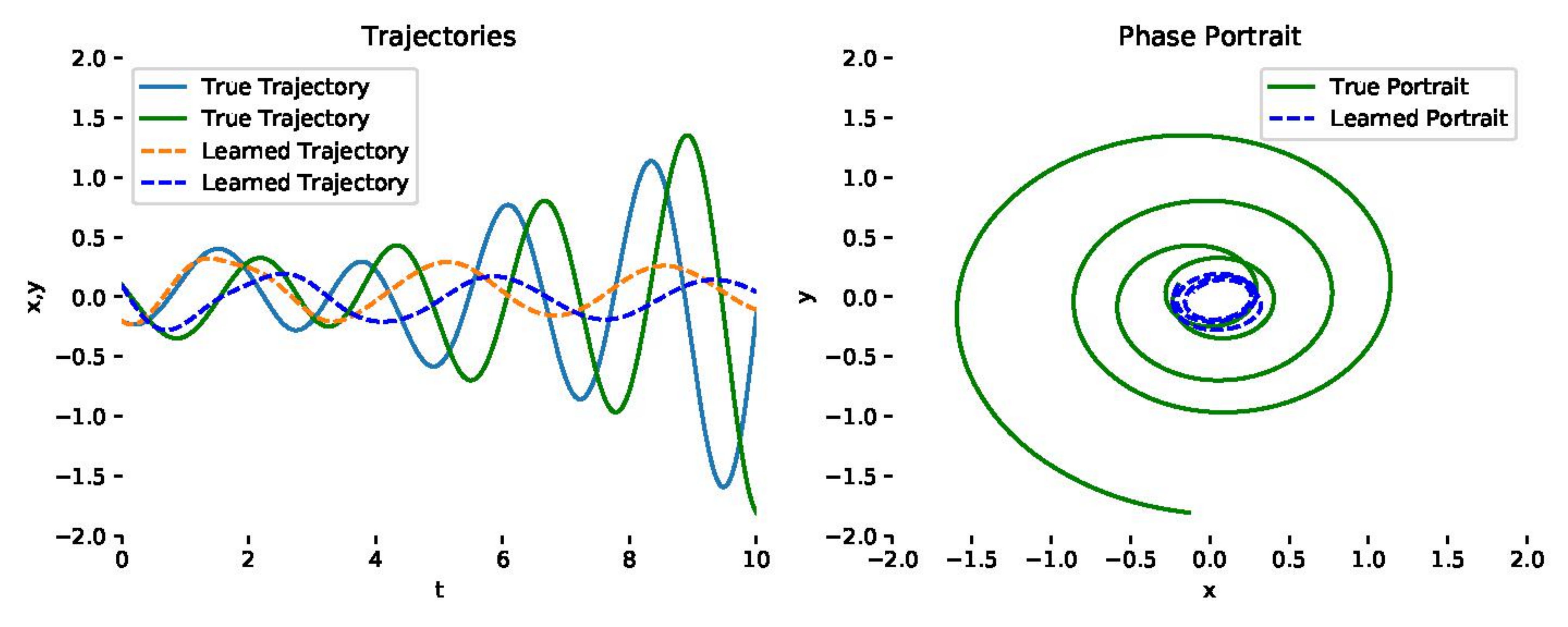}}
\subfigure[DDNN with delay 1.75]{\includegraphics[scale=0.15]{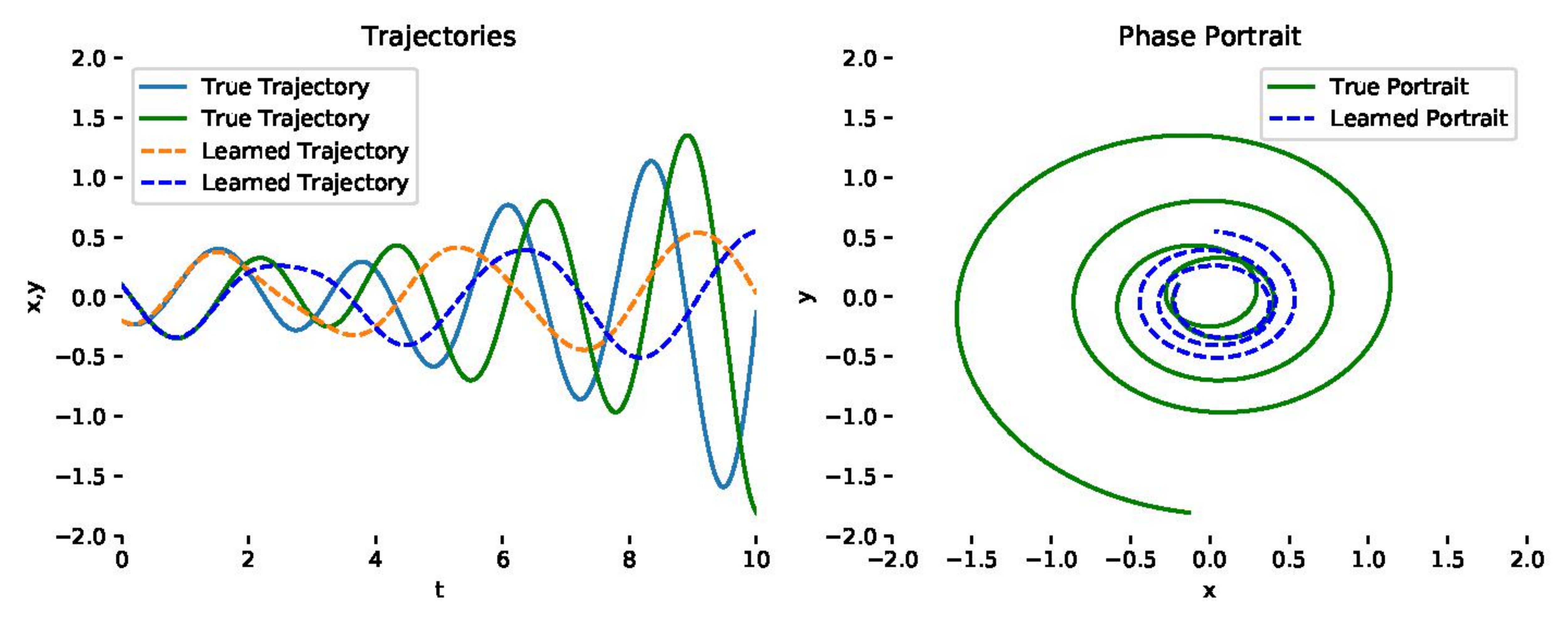}}
\subfigure[DDNN with delay 2.00]{\includegraphics[scale=0.15]{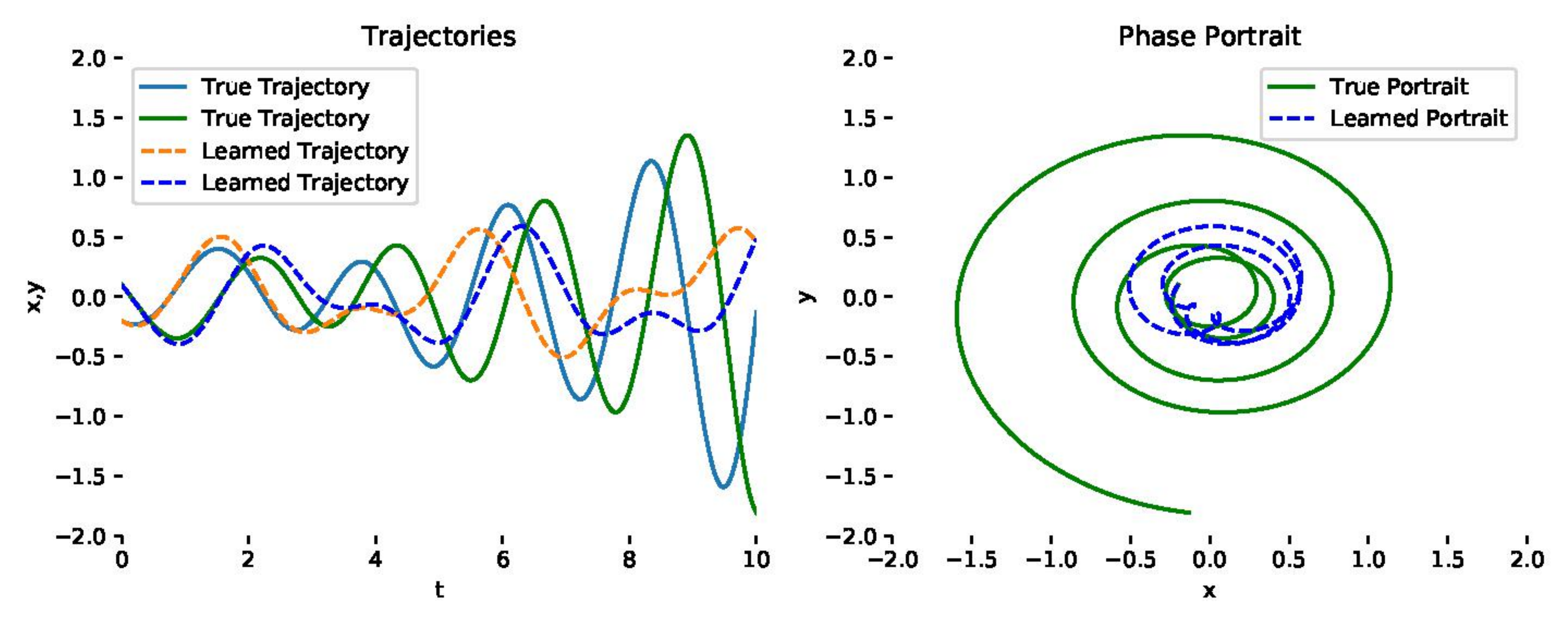}}
\subfigure[DDNN with delay 2.25]{\includegraphics[scale=0.15]{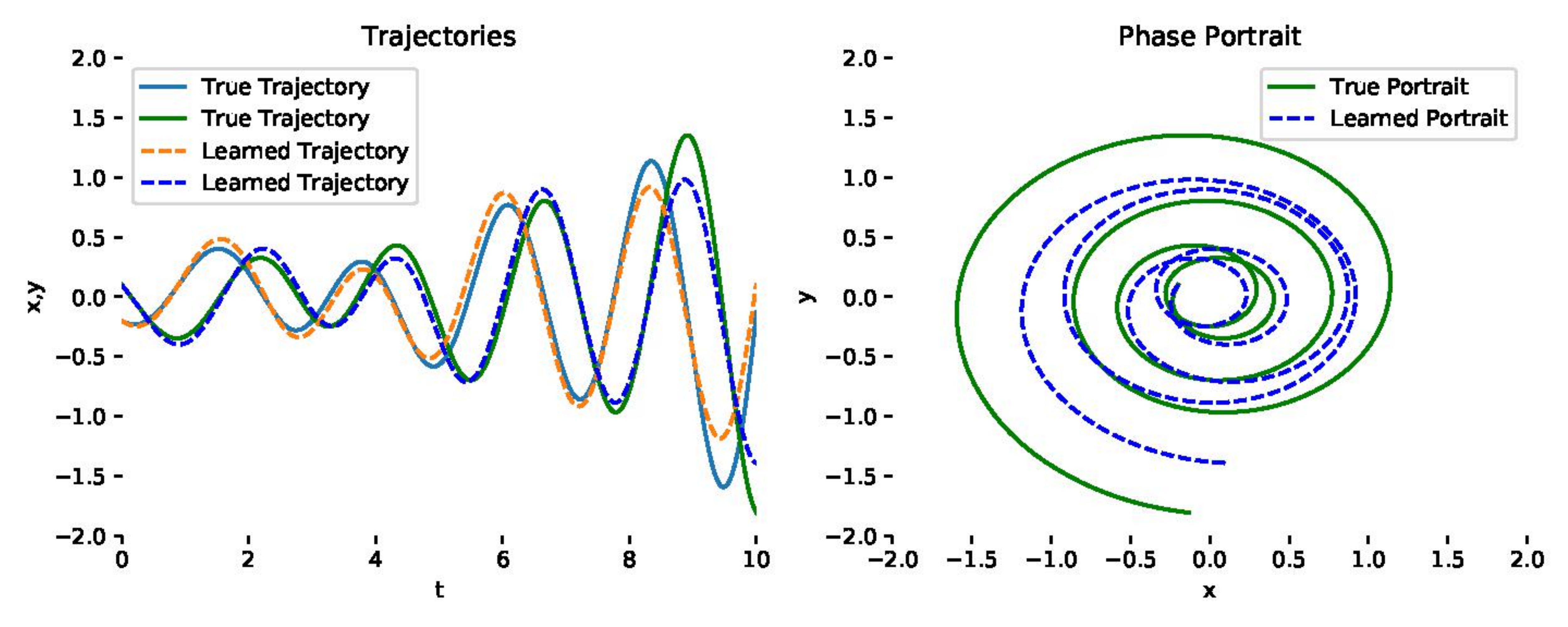}}
\subfigure[DDNN with delay 2.50]{\includegraphics[scale=0.15]{Images/delay_2.50.pdf}}
\subfigure[DDNN with delay 2.75]{\includegraphics[scale=0.15]{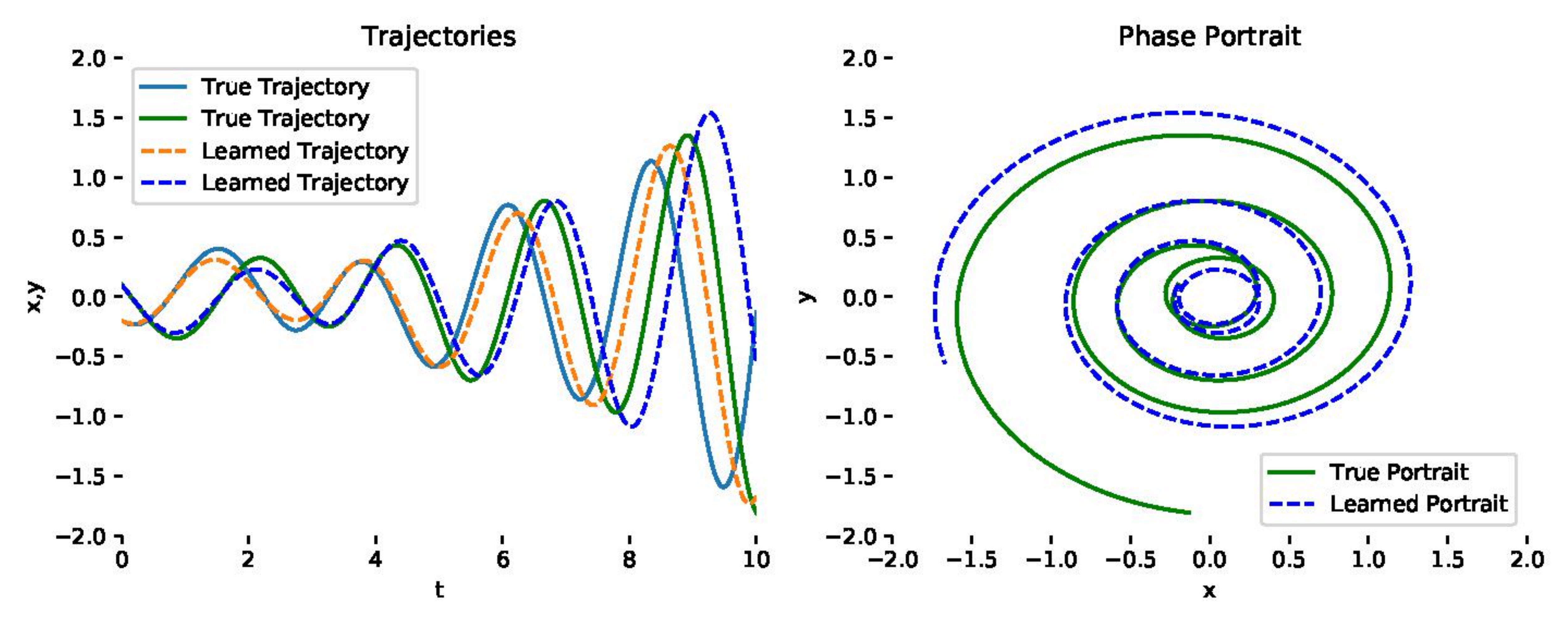}}
\subfigure[DDNN with delay 3.00]{\includegraphics[scale=0.15]{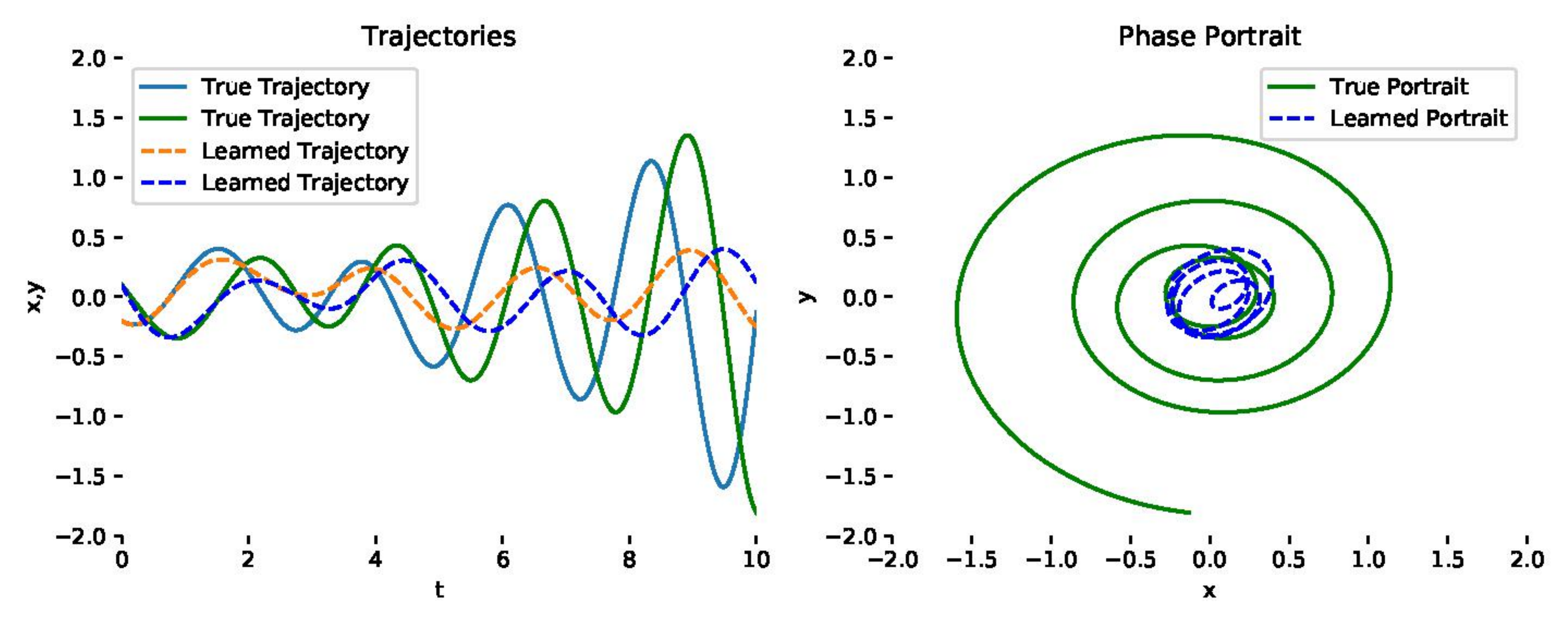}}
\subfigure[DDNN with delay 3.25]{\includegraphics[scale=0.15]{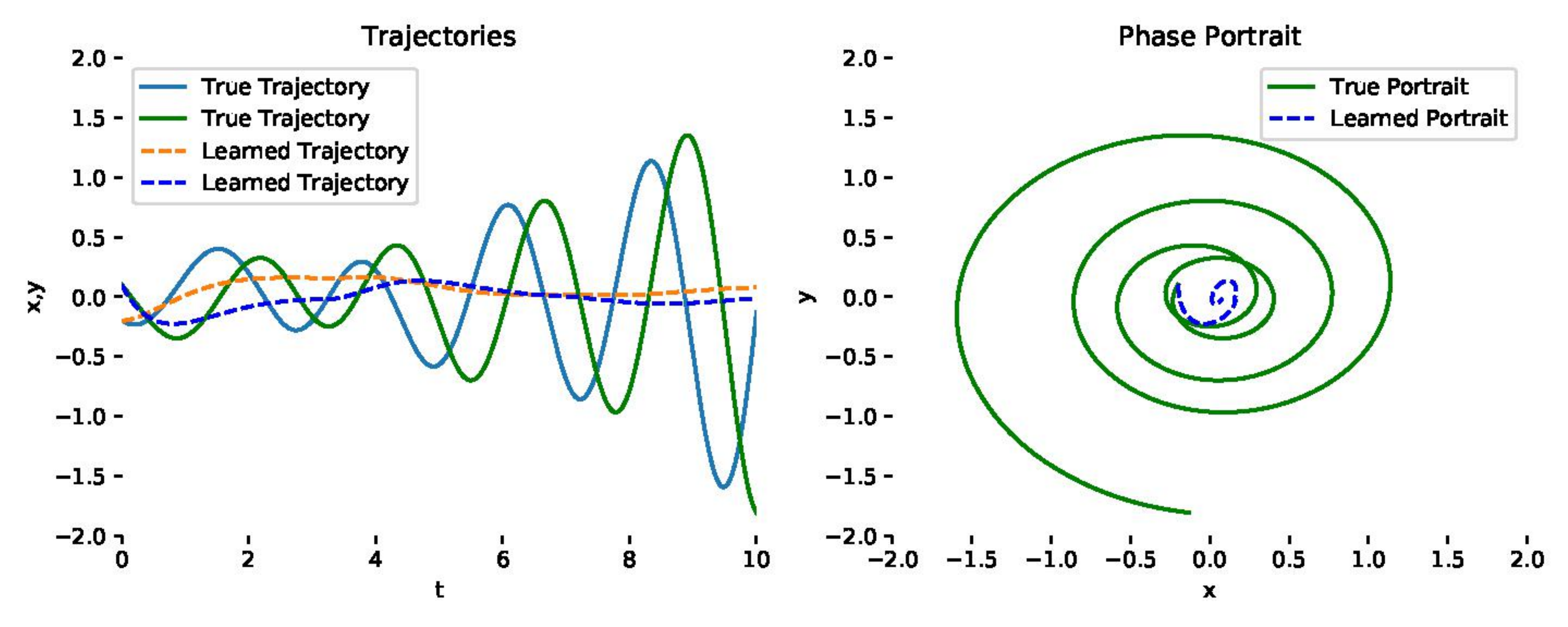}}
\subfigure[DDNN with delay 3.50]{\includegraphics[scale=0.15]{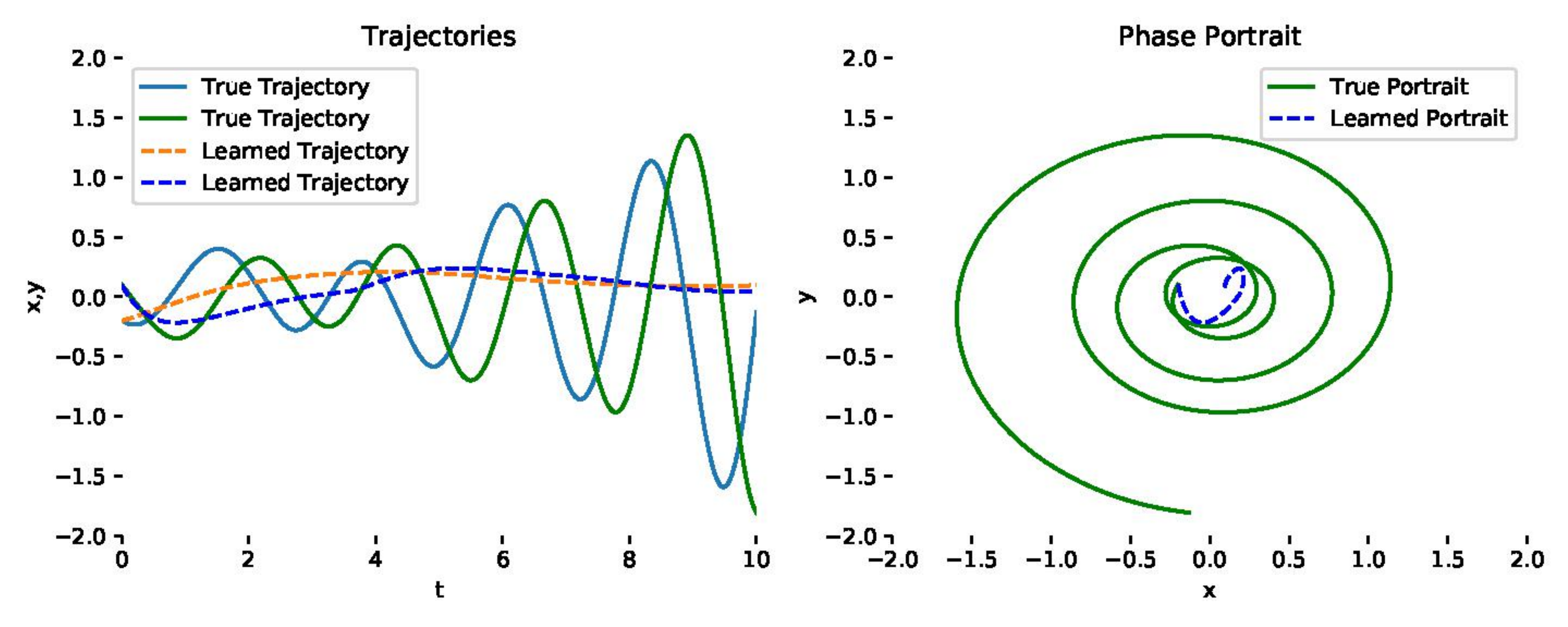}}

\end{center}
\caption{Showing the learning and extrapolating capabilities of DDNN models with different delay values, training samples are from the solution of DDE with delay 2.5.}
\label{fig:toy_data_delay}
\end{figure}

\end{appendix}
\end{document}